\documentclass[11pt,a4paper]{article}
\usepackage[hyperref]{emnlp2019}
\usepackage{times}
\usepackage{latexsym}
\usepackage{url}
\usepackage{amsmath}
\usepackage{multirow}
\usepackage{mathtools}
\usepackage{algorithm}
\usepackage{algpseudocode}

\aclfinalcopy
\title{Enhancing Context Modeling with a Query-Guided \\ Capsule Network for Document-level Translation}

\author{
  Zhengxin Yang$^1$$^2$$^3$, Jinchao Zhang$^3$, Fandong Meng$^3$ \\
  {\bf Shuhao Gu$^1$$^2$, Yang Feng$^1$$^2$$^\star$, Jie Zhou$^3$} \\
  $^{1}$ Key Laboratory of Intelligent Information Processing \\
  Institute of Computing Technology, Chinese Academy of Sciences (ICT/CAS) \\
  $^{2}$ University of Chinese Academy of Sciences \\
  $^{3}$ Pattern Recognition Center,	WeChat AI, Tencent Inc, China \\
  {\tt \{yangzhengxin17z,gushuhao17g,fengyang\}@ict.ac.cn} \\
  {\tt \{dayerzhang,fandongmeng,withtomzhou\}@tencent.com}
}

\date{}

\begin{document}
\maketitle
\newcommand\blfootnote[1]{%
\begingroup 
\renewcommand\thefootnote{}\footnote{#1}%
\addtocounter{footnote}{-1}%
\endgroup
}
\begin{abstract}
Context modeling is essential to generate coherent and consistent translation for Document-level Neural Machine Translations. The widely used method for document-level translation usually compresses the context information into a representation via hierarchical attention networks. However, this method neither considers the relationship between context words nor distinguishes the roles of context words. To address this problem, we propose a query-guided capsule networks to cluster context information into different perspectives from which the target translation may concern. Experiment results show that our method can significantly outperform strong baselines on multiple data sets of different domains.
\blfootnote{Joint work with Pattern Recognition Center, WeChat AI, Tencent Inc, China.}
\blfootnote{$\star$ Corresponding Author}
\end{abstract}

\section{Introduction}
The encoder-decoder based Neural Machine Translation (NMT) models \cite{sutskever-etal-2014-seq2seq,bahdanau-etal-2014n-eural,wu-etal-2016-google,vaswani-etal-2017-attention,zhang-etal-2019-bridging} have made great progresses and drawn much attention in recent years. In practical applications, NMT systems are often fed with a document-level input which requires reference resolution, the consistency of tenses and noun expressions and so on. Many researchers have proven that contextual information is essential to generate coherent and consistent translation for document-level translation \cite{hardmeier-2012-DiscourseIS,meyer-webber-2013-implicitation,sim-smith-2017-integrating,jean-etal-2017-DoesNM,maruf-haffari-2018-document,miculicich-etal-2018-document,zhang-etal-2018-improving,voita-etal-2018-ContextAwareNM,wang-etal-2017-exploiting-cross,tu-etal-2018-learning,maruf-etal-2019-selective}. Despite the great success of the above models, they are designed for sentence-level translation tasks and exclude contextual information in the model architecture.

On these grounds, the widely used Hierarchical Attention Networks (HAN) was proposed to integrate contextual information in document-level translation \cite{miculicich-etal-2018-document}. In this method, the context sentences are considered in the form of hierarchical attentions retrieved by the current generation. That is, it utilizes a word-level attention to represent a sentence and then a sentence-level attention to represent all the involved context. In this way, the final attention representation has to encode all the information needed for coherent and consistent translation, including reference information, tenses, expressions and so on. To get the multi-perspective information, it is necessary to distinguish the role of each context word and model their relationship especially when one context word could take on multiple roles \cite{zhang2018refining}. However, this is difficult to realize for the HAN model as its final representation for the context is produced with an isolated relevance with the query word which ignores relations with other context words.

To address the problem, we introduce Capsule Networks into document-level translation which have been proven good at modeling the \emph{parts-wholes} relations between lower-level capsules and higher-level capsules \cite{hinton-etal-2011-transforming,xiao-etal-2018-mcapsnet,sabour-etal-2017-dynamic,hinton-etal-2018-matrix,gu2019improving}. With capsule networks, the words in a context source sentence is taken as lower-level capsules and the information of different perspectives is treated as higher-level capsules. Then in the dynamic routing process of capsule networks, all the lower-level capsules trade off against each other and consider over all the higher-level capsules and drop themselves at a proper proportion to the higher-level capsules. In this way, the relation among lower-level capsules and that between lower-level capsules and higher-level capsules are both explored. In order to make sure higher-level capsules indeed cluster information needed by the target translation, we apply capsule networks to both sides of the current sentence and add a regularization layer using Pearson Correlation Coefficients (PCCs) to force the higher-level capsules on the two sides to approach to each other. In addition, we still need to ensure the final output of capsule networks is relevant to the current sentence. Therefore we propose a Query-guided Capsule Network (QCN) to have the current source sentence to take part in the routing process so that higher-level capsules can retain information related to the current source sentence.

To the best of our knowledge, this is the first work which applies capsule networks to document-level translation tasks and QCN is also the first attempt to customize attention for capsule networks in translation tasks. We conducted experiments on three English-German translation data sets in different domains and the results demonstrate that our method can significantly improve the performance of document-level translation compared with strong baselines.

\begin{figure*}
\centering
\includegraphics[width=0.6 \textheight]{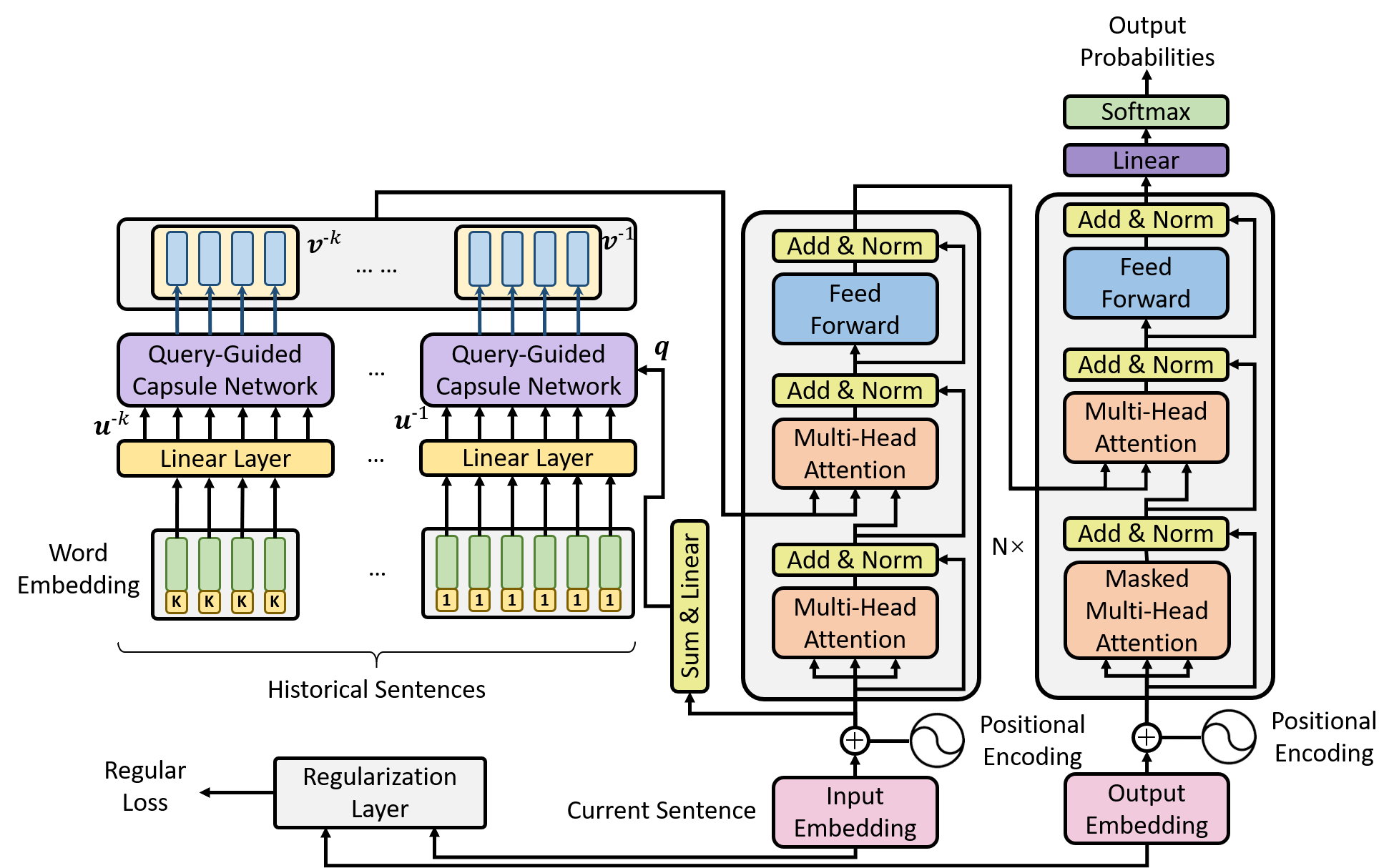}
\caption{\label{overall_arch}
The overall architecture consists of three modules: the \emph{Query-guided Capsule Network} in the upper left of this figure, the \emph{Regularization Layer} in the lower left of this figure and the \emph{Sub-layer-expanded Transformer} in the right of this figure.
}
\end{figure*}

\section{Background}
\subsection{Sentence-level NMT and Transformer Model}
Sentence-level NMTs are generally based on an encoder-decoder framework \cite{sutskever-etal-2014-seq2seq,bahdanau-etal-2014n-eural,wu-etal-2016-google,vaswani-etal-2017-attention,zhang-etal-2019-bridging,meng-zhang-2019-DTMT}. In this framework, the encoder encodes the source sentence \begin{math}\boldsymbol{x} = \{x_1, x_2, \cdots , x_M\}\end{math} into a sequence of continuous representations \begin{math}\boldsymbol{z} = \{z_1, z_2, \cdots, z_M\}\end{math}. Given \begin{math}\boldsymbol{z}\end{math}, the decoder predicts target words in order and returns the target translation \begin{math}\boldsymbol{y} = \{y_1, y_2, \cdots, y_N\}\end{math}. The objective function of NMT is to maximize the log-likelihood of a set of source-target language sentence pairs as
\begin{equation}\label{sent_trans_eq_1}
\mathcal{L}(\Theta;\boldsymbol{x}, \boldsymbol{y})
=
\frac{1}{|\mathcal{S}|} \sum_{(\boldsymbol{x}, \boldsymbol{y}) \in \mathcal{S}} \log \mathit{P}(\boldsymbol{y} | \boldsymbol{x} ; \Theta)
\end{equation}
where \begin{math}\mathcal{S}\end{math} is the training set.

As we implement our approaches based on the Transformer architecture \cite{vaswani-etal-2017-attention}, which is a strong sentence-level NMT baseline, we give a brief description of the Transformer model. The encoder and decoder are composed of similar layers which consists of two general types of sub-layers: multi-head attention mechanism and point-wise fully connected feed-forward network (FFN).

{\bf Encoder}: The encoder of the Transformer is composed of \begin{math}\mathit{N}\end{math} identical layers. Each of these layers includes a multi-head self-attention mechanism that allows each position of the output of previous encoder layer to attend to all other positions, as well as a position-wise FFN, which is stacked on top of the multi-head self-attention, composed of two linear transformations and a ReLU activation function.

{\bf Decoder}: The architecture of the decoder is similar to the encoder, however, it employs an additional multi-head attention sub-layer over the encoder output between the multi-head self-attention sub-layer and position-wise FFN sub-layer. The multi-head self-attention sub-layer needs to mask the input target tokens in the future.

\subsection{Document-level NMT}
The document-level translation task is to translate each source sentence with consideration of previous context in the document. Formally, the translation of a document \begin{math}\boldsymbol{D}\end{math} containing \begin{math}|\boldsymbol{D}| = \mathit{J}\end{math} sentence pairs can be defined as given the source document \begin{math}\boldsymbol{X}\end{math} in order and the translation system generates each translation \begin{math}\boldsymbol{y^{j}} \in \boldsymbol{Y}\end{math} in order. The document translation probability can be defined as:
\begin{equation}\label{docu_trans_eq_1}
\mathit{P}(\boldsymbol{Y} | \boldsymbol{X} ; \Theta)
=
\prod_{j=1}^{\mathit{J}} \mathit{P} (\boldsymbol{y}^{j}|\boldsymbol{x}^{j}, \boldsymbol{D}_{<j}  ; \Theta)
\end{equation}
where \begin{math}\boldsymbol{x}^{j}\end{math}, \begin{math}\boldsymbol{y}^{j}\end{math} denote the \begin{math}\mathit{j_{th}}\end{math} source and target sentence respectively, and \begin{math}\boldsymbol{D}_{<j}\end{math} denotes all of the previous sentence pairs in document.
For each sentence \begin{math}\boldsymbol{x}^{j}\end{math}, each target word is generated according to the source representation and the generated target hypothesis, therefore the Eq. (\ref{docu_trans_eq_1}) can be formulated as:
\begin{equation}\label{docu_trans_eq_2}
\mathit{P}(\boldsymbol{Y} | \boldsymbol{X} ; \Theta)
=
\prod_{j=1}^{\mathit{J}} \prod_{i=1}^{\mathit{I^{j}}} \mathit{P} (\mathit{y_{i}^{j}}|,\boldsymbol{y}_{<i}^{j}, \boldsymbol{x}^{j}, \boldsymbol{D}_{<j} ; \Theta)
\end{equation}
where \begin{math}\mathit{y_{i}^{j}}\end{math} denotes the \begin{math}\mathit{i_{th}}\end{math} word of the \begin{math}\mathit{j_{th}}\end{math} translation \begin{math}\boldsymbol{y}^{j}\end{math} with the length as \begin{math}\mathit{I^{j}}\end{math} and \begin{math}\boldsymbol{y}_{<i}^{j}\end{math} denotes the generated target hypothesis.

The training objective of document-level NMTs is to maximize the log-likelihood of translations in document context as following:
\begin{equation}\label{docu_trans_eq_3}
\mathcal{L}(\Theta;\boldsymbol{X}, \boldsymbol{Y})
=
\frac{1}{|\mathcal{D}|} \sum_{(\boldsymbol{X}, \boldsymbol{Y}) \in \mathcal{D}} \log \mathit{P}(\boldsymbol{Y} | \boldsymbol{X} ; \Theta)
\end{equation}
where \begin{math}\mathcal{D}\end{math} is the training set of the DocNMT.

Compared with the sentence-level NMTs, the critical part of document-level NMTs is to effectively capture and utilize the related contextual information when translating the to-be-translated source sentence.

\section{Our Approach}
In this section, we introduce the proposed Query-guided Capsule Network (QCN) for enhancing the document-level NMT. First, we present the overall architecture of the network, and then we describe the QCN in detail.

\subsection{Overall Architecture}
We aim to enhance the document-level NMT performance through effectively capturing and employing the contextual features in each historical sentence that related to the current sentence. We integrating a novel Query-guided Capsule Network into the sentence-level Transformer-based NMT \cite{vaswani-etal-2017-attention} to capture the document-level contextual information for translating the current source sentence.

As shown in Figure \ref{overall_arch}, the overall architecture of our translation model is composed of three modules:
\begin{itemize}
  \item \textbf{The Query-guided Capsule Network} takes the to-be-translated source sentence as the query to guide the procedure of retrieving related and helpful contextual features from historical sentences with a novel dynamic routing algorithm.
  \item \textbf{Sub-layer-expanded Transformer} contains a new sub-layer that attending the contextual features extracted from the QCN to effectively utilize them for translation. 
  \item \textbf{Regularization Layer} contains two conventional Capsule Networks to unify the source sentence and the target sentence into an identical semantic space through computing an extra PCCs loss item at the training stage.
\end{itemize}

\subsection{Query-guided Capsule Network}
The Capsule Network (CapsNet) \cite{sabour-etal-2017-dynamic} was proposed to build parts-wholes relationships in the iterative routing procedure, which can be used to capture features in historical sentences from lower level to higher level. Capsules in the lower layer vote for those in the higher layer by aggregating their transformations with iteratively updated coupling coefficients. However, there exists an obvious drawback of  directly applying the Capsule Network  into the document-level NMT for capturing contextual features. The reason is that the CapsNet can only extract internal features without considering whether features are related to the to-be-translated source sentence.

\begin{figure}
\centering
\includegraphics[width=0.3 \textheight]{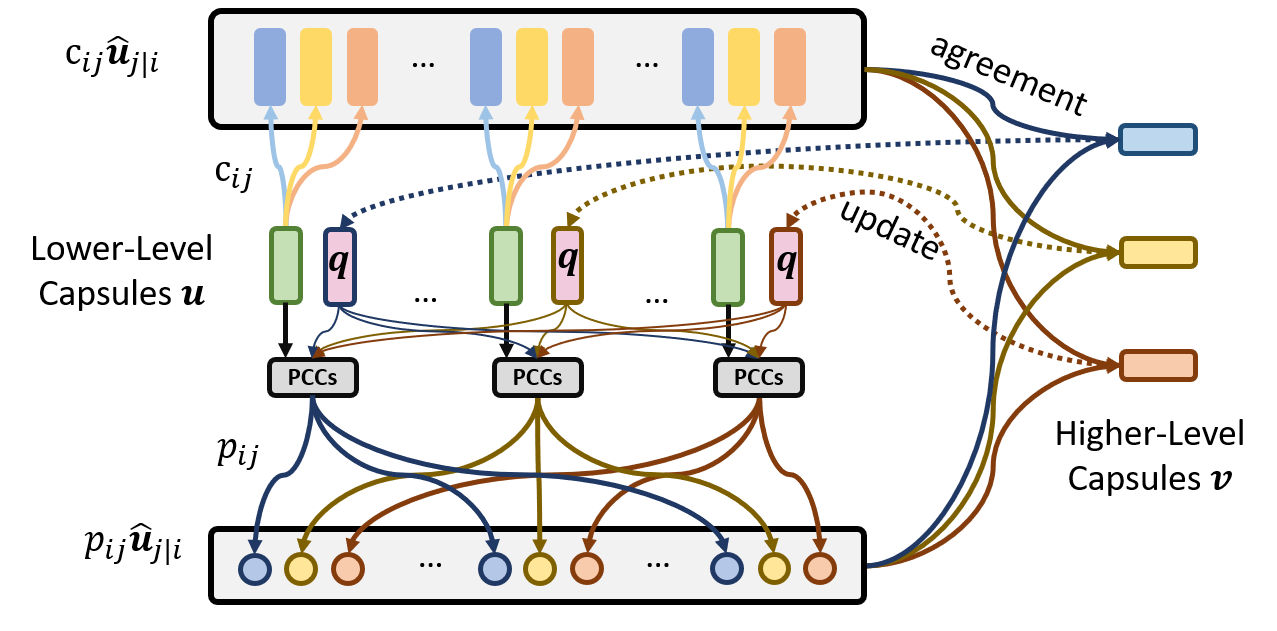}
\caption{\label{qcn_arch}
Query-guided Capsule Network, the green blocks in the middle of the figure are lower-level capsules \begin{math}\boldsymbol{u}_{i}\end{math}, right-side blocks indicate the higher-level capsules \begin{math}\boldsymbol{v}_{j}\end{math} and the query vector \begin{math}\boldsymbol{q}\end{math} is initially tiled to length \begin{math}|\boldsymbol{v}|\end{math} and updated with  corresponding higher-level capsules. Here shows a Query-guided Capsule Network with 3 high-level (feature) capsules. In each iteration, the PCCs is computed according to the input vector \begin{math}\boldsymbol{u}_{i}\end{math} and the query vector \begin{math}\boldsymbol{q}_{j}\end{math}, then \begin{math}p_{ij}\hat{\boldsymbol{u}}_{j|i}\end{math} and \begin{math}c_{ij}\hat{\boldsymbol{u}}_{j|i}\end{math} are added to obtain the higher-level capsules which will be used to update corresponding query later.
}
\end{figure}

To address this issue, we proposed the Query-guided Capsule Network (QCN), which employ the representation of the to-be-translated source sentence as the query to guide the feature extraction procedure of the Capsule Network. In this way, contextual features that generated from the QCN are based on the internal semantic relations among each historical sentence and the external semantic relations between historical sentences and the to-be-translated source sentence. The QCN is based on an improved dynamic routing algorithm which will be detailed introduced in the next section.

\subsubsection*{Improved Dynamic Routing of QCN}
\algrenewcommand{\algorithmicrequire}{\textbf{Input:}}
\algrenewcommand{\algorithmicensure}{\textbf{Output:}}
\begin{algorithm}[t]
\caption{Improved Dynamic Routing}\label{idr_alg}
\begin{algorithmic}[1]

{\small
  \Require{
    \begin{math}r\end{math}
    ,
    \begin{math}\boldsymbol{q}\end{math}
    and
    \begin{math}\boldsymbol{u} = \{\boldsymbol{u}_1, \boldsymbol{u}_2, \cdots, \boldsymbol{u}_{n}\end{math}\}
  }
  \Ensure{
    \begin{math}\boldsymbol{v} = \{\boldsymbol{v}_1, \boldsymbol{v}_2, \cdots, \boldsymbol{v}_{m}\}\end{math}
  } 
  \\
  // Initialization
  \For{ each output capsule \begin{math}\boldsymbol{v}_{j}\end{math} in higher-layer }
    \For{ each input capsule \begin{math}\boldsymbol{u}_{i}\end{math} in lower-layer }
      \State{\begin{math}\hat{\boldsymbol{u}}_{j|i} \gets \boldsymbol{W}_{ij} \boldsymbol{u}_{i}\end{math}}
      \State{\begin{math}\alpha_{ij} \gets 0\end{math}}
      \State{\begin{math}\boldsymbol{q}_{j} \gets \boldsymbol{q}\end{math}}
      \State{\begin{math} p_{ij} \gets {\rm tanh}({\rm PCCs}(\boldsymbol{u}_{i}, \boldsymbol{q}_{j})) \end{math}}
      \\
      \Comment{\begin{math}{\rm PCCs}\end{math} computes Eq \ref{pcc_eq}}
    \EndFor
  \EndFor
  \\
  \\

  // Iteration
  \State{\begin{math}itr \gets 0\end{math}}
  \Repeat
    \For{ each input capsule \begin{math}\boldsymbol{u}_{i}\end{math} in lower-layer}
      \State{\begin{math} \boldsymbol{c}_{i} \gets {\rm softmax}(\boldsymbol{\alpha}_{i}) \end{math}}
    \EndFor \\
    \For{ each output capsule \begin{math}\boldsymbol{v}_{j}\end{math} in higher-layer}
      \State{\begin{math} \boldsymbol{s}_{j} \gets \sum_{i=1}^{n} (c_{ij}+p_{ij}) \hat{\boldsymbol{u}}_{j|i} \end{math}}
      \State{\begin{math} \boldsymbol{v}_{j} \gets {\rm squash}(\boldsymbol{s}_{j}) \end{math}}
      \\
      \Comment{\begin{math}{\rm squash}\end{math} computes Eq \ref{squash_eq}}
    \EndFor \\

    \For{ each output capsule \begin{math}\boldsymbol{v}_{j}\end{math} in higher-layer }
      \For{ each input capsule \begin{math}\boldsymbol{u}_{i}\end{math} in lower-layer }
        \State{\begin{math}\alpha_{ij} \gets \alpha_{ij} + p_{ij}\hat{\boldsymbol{u}}_{j|i}\boldsymbol{v}_{j}\end{math}}
      \EndFor
      
      \State{\begin{math}\boldsymbol{q}_{j} \gets \frac{\boldsymbol{q}_{j} + \boldsymbol{v}_{j}}{2}\end{math}}
      
      \For{ each input capsule \begin{math}\boldsymbol{u}_{i}\end{math} in lower-layer }
        \State{\begin{math} p_{ij} \gets {\rm tanh}({\rm PCCs}(\boldsymbol{u}_{i}, \boldsymbol{q}_{j})) \end{math}}
      \EndFor
    \EndFor
    \State{\begin{math}itr \gets itr + 1\end{math}}
  \Until{ \begin{math}itr = r\end{math} }
}
\end{algorithmic}
\end{algorithm}

Given a query vector \begin{math}\boldsymbol{q}\end{math} and a set of input capsules \begin{math}\boldsymbol{u}=\{ \boldsymbol{u}_1, \boldsymbol{u}_2, \cdots, \boldsymbol{u}_{n}\}\end{math}, the dynamic routing algorithm iteratively calculates the correlation between the query vector and each input capsule and updates output capsules. Specifically, query vector \begin{math}\boldsymbol{q}\end{math} is the representation of source to-be-translated sentence and input capsules \begin{math}\boldsymbol{u}\end{math} are all word embeddings in previous sentences, which can be formulated as:
\begin{gather}
\boldsymbol{q} = {\rm g}(\sum_{x \in \boldsymbol{x}}{\rm embedding}(x))\\
\boldsymbol{u}_{i}^{-k} = {\rm f}([{\rm embedding}(x_{i}^{-k}); {\rm onehot}(k)])
\end{gather}
where \begin{math}f\end{math} and \begin{math}g\end{math} are both linear transformation functions and \begin{math}k\end{math} indicates the distance of historical sentence from the to-be-translated sentence. Each word embedding in historical sentences is concatenated with a distance-determined one-hot vector to provide positional markers. Compared to the dynamic routing method proposed by \cite{sabour-etal-2017-dynamic}, our improved dynamic routing method of QCN can model information that related to the query among input capsules.

Algorithm \ref{idr_alg} shows details of the algorithm and Figure \ref{qcn_arch} illustrates the interaction of the various components in the QCN. Initially, the improved dynamic routing algorithm of QCN recieves a sequence of lower-level capsules \begin{math}\boldsymbol{u} = \{ \boldsymbol{u}_1, \boldsymbol{u}_2, \cdots, \boldsymbol{u}_{n}\}\end{math} and a query vector \begin{math}\boldsymbol{q}\end{math} and then calculates a Pearson Correlation Coefficients (PCCs) between \begin{math}\boldsymbol{q}\end{math} and each input capsule \begin{math}\boldsymbol{u}_{i}\end{math} (line 7). The PCCs is a measure of the linear correlation between two variables. When PCCs is close to +1, it means they have very strong positive linear correlation, and close to -1 means total negative correlation. Given a pair of variables \begin{math}(\boldsymbol{A}, \boldsymbol{B})\end{math}, the formula for computing PCCs is
\begin{multline}\label{pcc_eq}
{\rm PCCs}(\boldsymbol{A}, \boldsymbol{B}) 
= \frac{{\rm Cov}(\boldsymbol{A}, \boldsymbol{B})}{\sigma_{\boldsymbol{A}}\sigma_{\boldsymbol{B}}}\\
= \frac{(\boldsymbol{A}-\frac{1}{n}\sum_{i=1}^{n}a_{i})^T(\boldsymbol{B}-\frac{1}{n}\sum_{i=1}^{n}b_{i})}{\|\boldsymbol{A}-\frac{1}{n}\sum_{i=1}^{n}a_{i}\|\cdot\|\boldsymbol{B}-\frac{1}{n}\sum_{i=1}^{n}b_{i}\|}
\end{multline}
where \begin{math}\boldsymbol{A}\end{math} and \begin{math}\boldsymbol{B}\end{math} are both \begin{math}n\end{math}-dimension vectors, \begin{math}{\rm Cov}\end{math} is the covariance and \begin{math}\sigma_{\boldsymbol{A}}, \sigma_{\boldsymbol{B}}\end{math} is the standard deviation of \begin{math}\boldsymbol{A}\end{math} and \begin{math}\boldsymbol{B}\end{math} respectively.

The routing iteration process then computes coupling coefficients, denoted as \begin{math}\boldsymbol{c}_{i}\end{math}(line 16), with regard to a input capsule \begin{math}\boldsymbol{u}_{i}\end{math} and all the higher-level capsules \begin{math}\boldsymbol{v}\end{math}. In the original dynamic routing algorithm \cite{sabour-etal-2017-dynamic}, coupling coefficients are only determined by the cumulative ``agreement'', which are the prior probabilities that capsule \begin{math}\boldsymbol{u}_{i}\end{math} should be coupled to capsule \begin{math}\boldsymbol{v}_{j}\end{math}. However, in DocNMT situation, it is far from enough to cluster the high-level information from the capsules in lower-level that related to query vector according to a naive ``agreement''. To address this issue, we reduce the ``agreement'', when PCCs show a negative linear correlation between the query vector \begin{math}\boldsymbol{q}\end{math} and the input vector \begin{math} \boldsymbol{u}_{i} \end{math}. Instead, we increase the ``agreement'', when PCCs are positive (line 27). The query vector \begin{math}\boldsymbol{q}\end{math} is initially tiled to length \begin{math}|\boldsymbol{v}|\end{math}  and updated with the corresponding higher-level capsules \begin{math}\boldsymbol{v}_{j}\end{math} in each iteration (line 29).

Our routing iteration updates higher-level capsules by adding \begin{math} \sum_{i=1}^{n} p_{ij} \hat{\boldsymbol{u}}_{j|i} \end{math} to \begin{math} \boldsymbol{s}_{j} \end{math}. This step can add more information related to the query vector and cut off the unrelated features (line 20). It is necessary to get different length of output capsules shrunk into the 0 to 1 interval using ``squash'' function (line 21) proposed by \citet{sabour-etal-2017-dynamic} which is shown in Eq.(\ref{squash_eq}).
\begin{equation}\label{squash_eq}
{\rm squash}(\boldsymbol{t}) = \frac{||\boldsymbol{t}||^2}{1+||\boldsymbol{t}||^2} \frac{\boldsymbol{t}}{||\boldsymbol{t}||}
\end{equation}
where \begin{math}\boldsymbol{t}\end{math} can be the initial input capsule \begin{math}\boldsymbol{u}_{i}\end{math} or the vector \begin{math}\boldsymbol{s}_{j}\end{math} for predicting the output capsule.

\subsection{Sub-layer-expanded Transformer}
To effectively utilize contextual features extracted from each historical sentence by QCN, we introduce an additional context-aware multi-head attention sub-layer in each layer of the Transformer encoder. The multi-head attention can attend to all the positions of the contextual features with outputs of the previous sub-layer as a query. Then the output of the context-aware multi-head attention sub-layer is fed into the point-wise feed-forward sub-layer in each layer of the encoder. The right part in Figure \ref{overall_arch} shows details of the Sub-layer-expanded Transformer. Specifically, the equation of multi-head attention computing procedure is as following:
\begin{multline}\label{multihead_eq}
{\rm MultiHead}(\!Q,\!K,\!V)\!=\!{\rm Concat}(H_1,\!\cdots\!,H_h)W^O \\
H_i = {\rm Attention}(Q{W_i}^Q, K{W_i}^K, V{W_i}^V)
\end{multline}
where \begin{math}{W_i}^Q\end{math}, \begin{math}{W_i}^K\end{math} and \begin{math}{W_i}^V\end{math} are the parameter matrices, \begin{math}Q\end{math}, \begin{math}K\end{math} and \begin{math}V\end{math} indicates the query, key and value representations. The computation of the attention function is as following:
\begin{equation}\label{attn_eq}
{\rm Attention}(Q,K,V) = {\rm softmax}(\frac{QK^T}{\sqrt{d_k}})V
\end{equation}
where \begin{math}d_k\end{math} indicates the dimension of queries \begin{math}Q\end{math} and keys \begin{math}K\end{math}.

\subsection{Regularization Layer}
\begin{figure}
\centering
\includegraphics[width=0.32 \textheight]{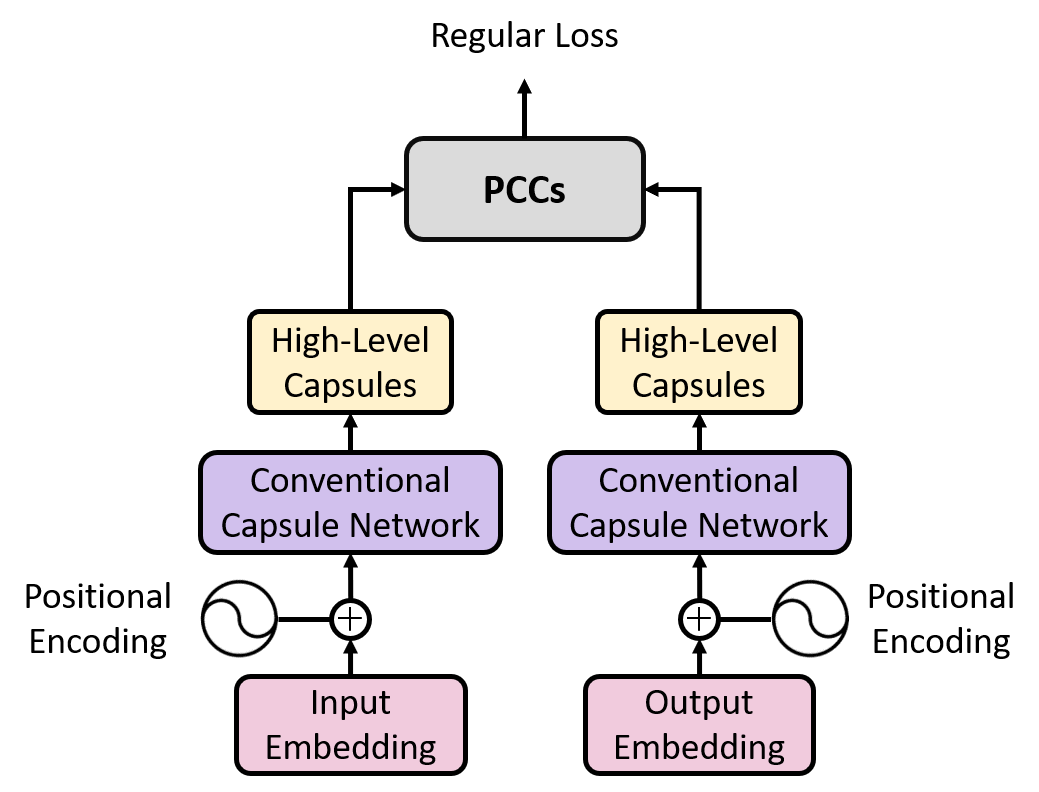}
\caption{\label{reg_arch}
Regularization Layer. In the training stage, this layer takes the inputs of the encoder (left part of the figure) and decoder (right part of the figure) as the input capsules of two capsule networks separately, and computes the PCCs between the high-level capsules of two networks.
}
\end{figure}
To better modeling the document-level translation task, we incorporate a regularization layer (Figure \ref{reg_arch}) into the whole architecture to restrict the source sentence and target sentence to an identical semantic space. This layer separately feeds the inputs of encoder and decoder into two capsule networks \begin{math}{\rm Caps}_{enc}\end{math} and \begin{math}{\rm Caps}_{dec}\end{math}, and computes the PCCs between the outputs of two networks. We regard the PCCs as an extra regularization term in the final objective at the training stage. The loss function of our model can be formulated as:
\begin{multline}\label{loss_eq}
\mathcal{L}(\Theta;\boldsymbol{X}, \boldsymbol{Y})
=
\frac{1}{|\mathcal{D}|}\\
\times \sum_{j=1}^{\mathit{J}} \sum_{i=1}^{\mathit{I^{j}}} \{\log \mathit{P} (\mathit{y_{i}^{j}}|,\boldsymbol{y}_{<i}^{j}, \boldsymbol{x}^{j}, \boldsymbol{D}_{<j} ; \Theta)\\
+ {\rm PCCs}({\rm Caps}_{enc}(\boldsymbol{x}^{j}), {\rm Caps}_{dec}(\boldsymbol{y}^{j}))\} \\
\end{multline}
where \begin{math}\Theta\end{math} are parameters of the model, \begin{math}\boldsymbol{D}_{<j}\end{math} are historical sentences of the to-be-translated source sentence, \begin{math}\boldsymbol{x}^{j}\end{math} is the to-be-translated sentence and \begin{math}\boldsymbol{y}_{<i}^{j}\end{math} denotes the generated target hypothesis.

\section{Experiments}
\subsection{Settings}
\begin{table*}
\centering
  \begin{tabular}{l | c c | c c | c c}
              & \multicolumn{2}{|c|}{\textbf{TED}} & \multicolumn{2}{|c|}{\textbf{News}} & \multicolumn{2}{|c}{\textbf{Europarl}} \\
              &  Sent No. & Doc len avg &  Sent No.  & Doc len avg &  Sent No.  &  Doc len avg \\
  \hline
  \hline
  Training    &  206,126  & 121.39      &  236,287  &  38.93       &  1,666,904 &  14.14       \\
  Development &  8,967    & 96.42       &  2,169    &  26.78       &  3,587     &  14.95       \\
  Test        &  2,271    & 98.74       &  2,999    &  19.35       &  5,134     &  14.26       \\
  \end{tabular}
\caption{\label{corpora_table}
The statistics of the training/development/test corpora in number of sentence pairs and the average document length (in sentences).
}
\end{table*}

\subsubsection*{Datasets and Evaluation Metrics}
We carry out experiments on English-German translation tasks in three different domains: talks, news, and speeches. The corpora statistics are shown in Table \ref{corpora_table}.

\begin{itemize}
  \item
  {\bf TED.} This corpus is a Machine Translation part of the IWSLT 2017 \cite{cettolo-etal-2012-EAMT2012} evaluation compaigns\footnote{https://wit3.fbk.eu/}, each TED talk is considered to be a document. we take the {\it tst2016-2017} as the test set, and other as our development set.

  \item
  {\bf News.} We take the sentence-aligned document-delimited News Commentary v11 corpus\footnote{http://www.casmacat.eu/corpus/news-commentary.html} as our training set. The WMT'16 {\it news-test2015} and {\it news-test2016} are used for development and testing respectively.

  \item
  {\bf Europarl.} The corpus are extracted from the Europarl v7 \cite{koehn-2005-europarl} according to the method mentioned in \citet{maruf-haffari-2018-document}. The training, development and test sets are obtained through randomly splitting the corpus.
\end{itemize}

We download all of above extracted corpora\footnote{https://github.com/sameenmaruf/selective-attn/tree/master/data} from \citet{maruf-etal-2019-selective}. The tokenization and truecase pre-processing are implemented on all datasets using the scripts of the Moses Toolkit\footnote{https://github.com/moses-smt/mosesdecoder/tree/master/scripts} \cite{koehn-etal-2007-moses}. We also apply segmentation into BPE subword units\footnote{https://pypi.org/project/subword-nmt/} \cite{sennrich-etal-2016-neural} with 30K merge operations.

We use two metrics: BLEU \cite{papineni-etal-2002-bleu} and Meteor \cite{lavie-agarwal-200-7meteor} to evaluate the translation quality.

\subsubsection*{Models and Baselines}
We use the Transformer architecture as our context-agnostic baseline and adopt three context-aware baselines \cite{zhang-etal-2018-improving,miculicich-etal-2018-document,maruf-haffari-2018-document}.

We performed the same configuration on our models\footnote{The code is available at https://github.com/Jason-Young-NLP/DocNMT-Transformer} according to the settings of the \citet{maruf-haffari-2018-document}. Specifically, for the Transformer, we set the hidden size and point-wise FFN size as 512 and 2048 respectively. We use 4 layers and 8 attention heads in both encoder and decoder. All dropout rates are set to 0.1 for context-agnostic model and 0.2 for context-aware model.

In the training phase, we use the default Adam optimizer \cite{kingma-ba-2014-adam} with a fixed learning rate of 0.0001. The batch size is 1500 on TED dataset and 900 on both News and Europarl datasets.

\subsection{Results and Analysis}
\begin{table*}
\centering
\scalebox{0.95}{
    \begin{tabular}{l | c c | c c | c c }
           & \multicolumn{2}{c|}{\textbf{TED}} & \multicolumn{2}{c|}{\textbf{News}} & \multicolumn{2}{c}{\textbf{Europarl}} \\
       \textbf{Model}                                  &  \textbf{BLEU} & \textbf{Meteor}  &  \textbf{BLEU} & \textbf{Meteor}   &    \textbf{BLEU} & \textbf{Meteor}    \\
       \hline
       \hline
       Transformer-DocNMT \cite{zhang-etal-2018-improving}      & 24.00 &  44.69  & 23.08 & 42.40    &  29.32  & 46.72     \\
       HAN-DocNMT \cite{miculicich-etal-2018-document}  & 24.58 &  45.48  & \textbf{25.03} & \textbf{44.02}    &  28.60  & 46.09     \\
       SAN-DocNMT \cite{maruf-etal-2019-selective}      & 24.42 &  45.38  & 24.84 & 44.27    &  29.75  & 47.22     \\
       \hline
       Transformer & 23.28 &  44.17  &  21.67 & 41.11    &  28.72  & 46.22    \\
      
       Transformer + Regularization Term                         & 24.55 &  45.57   &  22.09   & 41.77   &   29.42 &  47.59       \\
       Transformer + QCN                                  & 24.41 &  \textbf{46.09}   &   22.22   & 41.90   &    29.48  &  47.49      \\
       Transformer + QCN + Regularization Term                   & \textbf{25.19} &  45.91   &  22.37  & 41.88   &  \textbf{29.82} & \textbf{47.86} \\
    \end{tabular}
}
    \caption{\label{bleu_tab}
    BLEU and Meteor scores of models. There exists three context-aware baseline models and a context-agnostic model.  ``Regularization Term'' denotes integrating the regularization layer into the Transformer model and ``QCN'' denotes incorporating  Query-guided Capsule Network in the Transformer model.
    }
\end{table*}

\subsubsection*{Main Results}
Table \ref{bleu_tab} shows that our model surpasses all the context-agnostic \cite{vaswani-etal-2017-attention} and context-aware \cite{zhang-etal-2018-improving,miculicich-etal-2018-document,maruf-haffari-2018-document} baselines on TED and Europarl datasets. For TED dataset, the performance of our model greatly exceeds that of all other baselines, and is better than \citet{miculicich-etal-2018-document} with a gain of +0.59 BLEU and +0.61 Meteor. For Europarl dataset, our model got improvements with a gain of +0.07 on BLEU metric, but the Meteor score is +0.64 higher than \citet{maruf-etal-2019-selective} which utilize the whole document as the contextual information, whereas we only using 3 previous sentences. 

Results on the sequence-level Transformer and our DocNMTs show that the captured contextual features provide helpful semantic information for enhancing the translation quality. The regularization term that we proposed can effectively further improve the model performance on TED and Europarl datasets. For the restriction of the GPU memory, we have to filter long sentences to keep our model running. Although, it hurts the model performance on the ``News'' dataset (contains many long sentences), the QCN module and regularization term still bring improvements. 

\subsubsection*{Effect of Contextual Information Scope}
\begin{figure}
\centering
\includegraphics[width=0.3 \textheight]{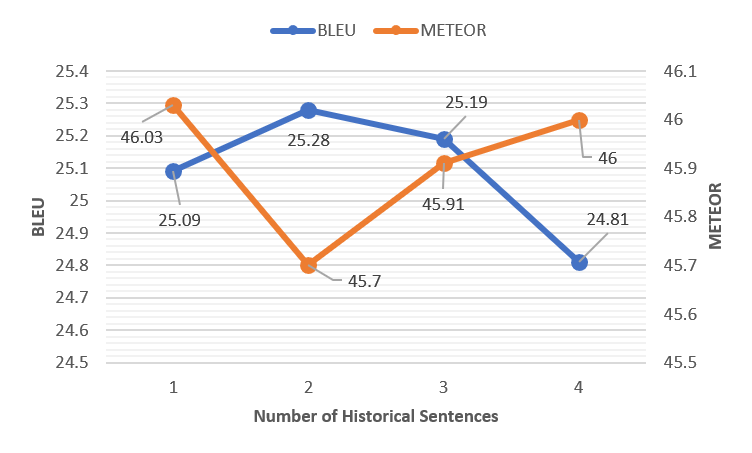}
\caption{\label{noh_chart}
Effect of Contextual Information Scope. It is not that the more historical sentences we utilize, the better translation performance is. Effect of scope on BLEU and Meteor scores are inconsistent to some degree.
}
\end{figure}

To investigate the effect of contextual information scope, we carry on the number of historical sentences hyper-parameter experiments on the TED talk dataset. We fix the hyper-parameters of the QCN by setting both the number of higher-level capsules and routing iteration to four, and investigating the impact of changes in the number of historical sentences on BLEU and Meteor scores. Figure \ref{noh_chart} shows that using one historical sentence in QCN can obtain the best Meteor score while the highest BLEU score is presented when we utilize two historical sentences. We found that the growth of Meteor and BLEU scores are opposite. Therefore, the choice of utilizing how many historical sentences is a trade-off between both scores. We choose three historical sentences as our final setting. Through experimentation, we also found it is not that the more historical sentences we utilize, the better translation performance is.

\begin{figure}
\centering
\includegraphics[width=0.3 \textheight]{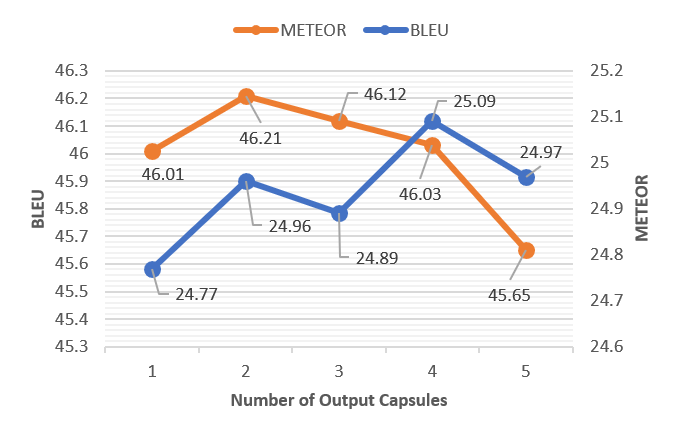}
\caption{\label{noc_chart}
Effect of Feature Capsule Number. Appropriate capsule number is important for model performance.
}
\end{figure}

\subsubsection*{Effect of Feature Capsule Number}
QCN is the crucial part of our overall architecture and the positive and negative impact depends on the configuration of the QCN. Therefore, We also investigate the effect of the hyper-parameter of QCN: the number of feature capsules.

We set the number of historical sentence as 3 according to the previously experimental results, and the number of routing iteration is set to 4. Figure \ref{noc_chart} shows that Meteor score become highest when the number of higher-level capsules is set as 2, but BLEU score can obtain best score at 4. We finally choose 4 as the final setting because both BLEU and Meteor can obtain relatively good results. 

\begin{figure}
\centering
\includegraphics[width=0.3 \textheight]{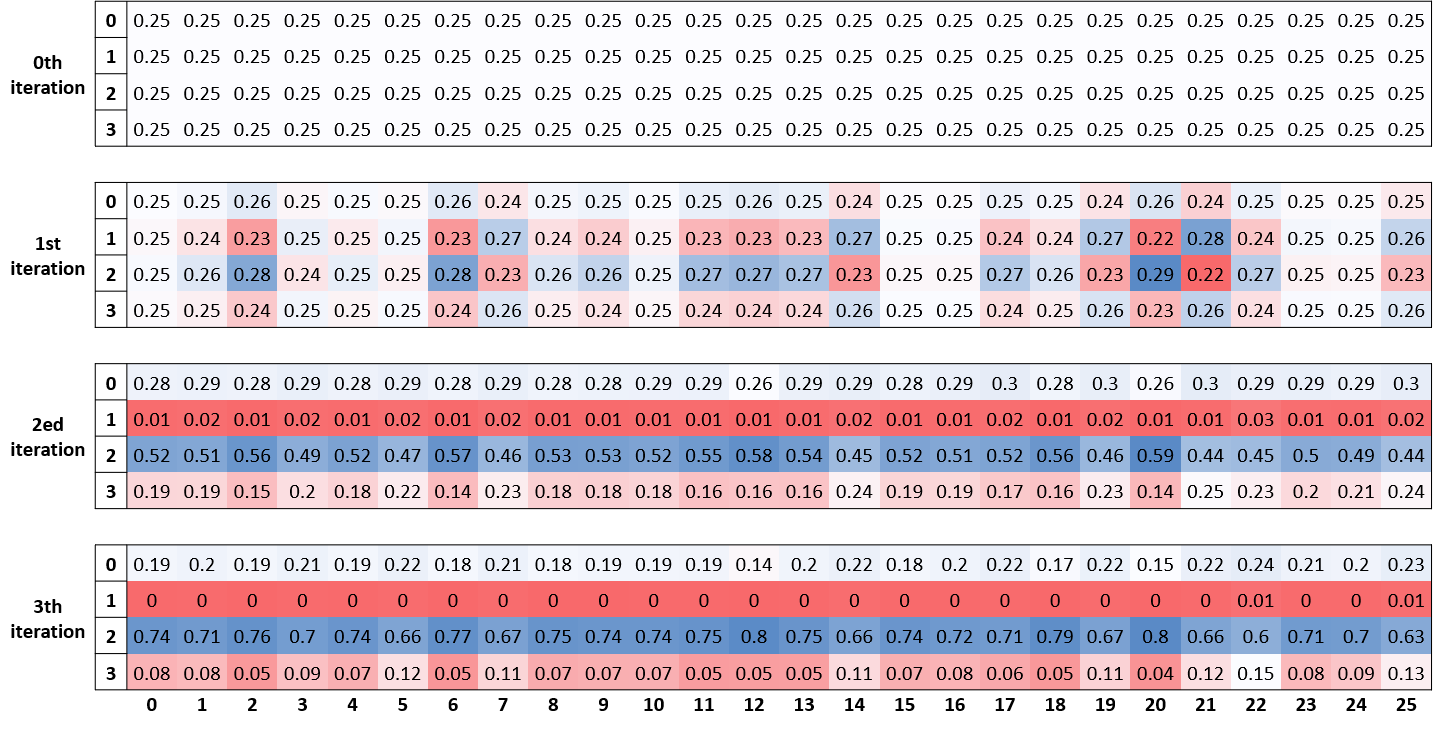}
\caption{\label{ccs_heat}
Agreement distribution of a sentence (26 words) as inputs of the QCN and 4 contextual features as output. Blue color means higher agreement and the red means lower. From top to bottom are four heat maps in \begin{math}0_{th}\end{math} to \begin{math}3_{rd}\end{math} iterations.
}
\end{figure}

\subsubsection*{Visualization of Agreement and PCCs}
Coupling Coefficients (CCs) can indirectly reflect the variation of the ``agreement'', so we visualize the coefficients in each routing iteration at the stage of decoding as shown in Figure \ref{ccs_heat}. In the first iteration, all coupling coefficients are initialized in a uniform distribution, all higher-level capsules are voted by lower-level capsules equally. Then, lower-level capsules are iteratively trained to send more information to the proper higher-level capsule. Figure \ref{ccs_heat} shows that most of input capsules are tend to vote the \begin{math}2_{ed}\end{math} feature capsule finally.

\begin{figure}
\centering
\includegraphics[width=0.3 \textheight]{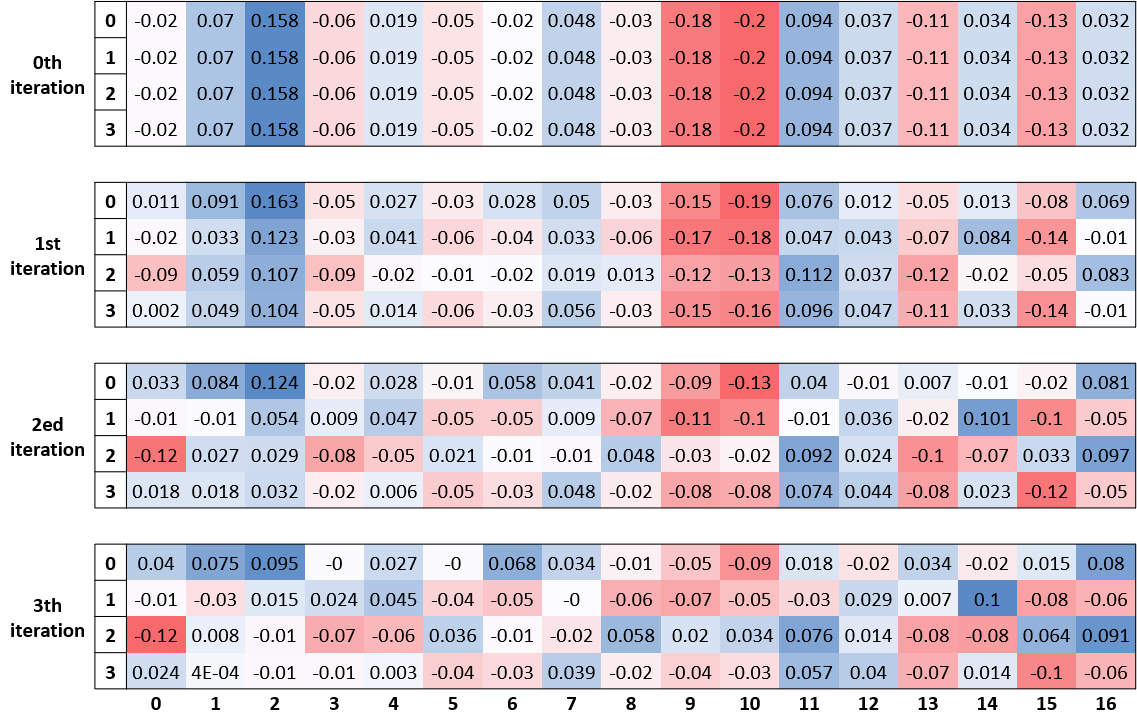}
\caption{\label{pccs_heat}
PCCs distribution of a sentence has 17 words as inputs of the QCN and 4 duplicated query. Blue color denotes positive correlation and red means negative. From top to bottom are four heat maps in \begin{math}0_{th}\end{math} to \begin{math}3_{rd}\end{math} iterations.
}
\end{figure}

Different from the CCs, PCCs show the linear correlation between query and inputs. Initially, the query vector is tiled with the number of feature capsules and is used to calculate the PCCs with each input capsules, as the Figure \ref{pccs_heat} shows that the color of every column is identical. As the iterative routing begins, each query is updated according to the higher-level capsules. The function of the PCCs is to increase or decrease the coupling coefficients value according to the positive or negative value. See Figure \ref{pccs_heat}, we can find that PCCs varies as iteration changes.

\section{Related Work}
\subsection*{Document-level Machine Translation}
Document-level machine translation became a hot research direction in the later stage of statistical machine translation era. \citet{hardmeier-federico-2010-modelling} represented the links between word pairs in the context using a word dependency model for SMT to improve the translation of anaphoric pronouns. \citet{hardmeier-etal-2012-document,hardmeier-etal-2013-docent} first proposed a new document-level SMT paradigm that translates whole documents as units. However, in this period, most of the work has not achieved too many compelling results or has been only focused on a part of difficulties.

With the coming of the era of Neural Machine Translation, many works began to focus on Document-level NMT tasks. \citet{xiong-etal-2019-ModelingCF} trained a reward teacher to refine the translation quality from a document perspective.  \citet{tiedemann-scherrer-2017-neural} simply concatenated sentences in one document as models' input or output. \citet{jean-etal-2017-DoesNM} used additional context encoder to capture larger-context information.  \citet{kuang-etal-2017-cache,tu-etal-2018-learning} used a cache to memorize most relevant words or features in previous sentences or translations.

Recently, several studies integrated additional modules into the Transformer-based NMTs for modeling contextual information \cite{voita-etal-2018-ContextAwareNM,zhang-etal-2018-improving}. \citet{maruf-haffari-2018-document} proposed a document-level NMT using a memory-networks, and \citet{wang-etal-2017-exploiting-cross} and \citet{miculicich-etal-2018-document} integrated hierarchical attention network in RNN-based NMT or Transformer to model the document-level information. \citet{maruf-etal-2019-selective} used the whole document as the contextual information and firslty divided document-level translation tasks into two types: offline and online.

\subsection*{Capsule Networks}
\citet{hinton-etal-2011-transforming} proposed the capsule conception to use vector for describing the pose of an object. The dynamic routing algorithm was proposed by \citet{sabour-etal-2017-dynamic} to build the part-whole relationship through the iterative routing procedure. \citet{hinton-etal-2018-matrix} designed a new routing style based on the EM algorithm. Some researchers investigated to apply the capsule network for various tasks. \citet{wang-etal-2019-towards} investigated a novel capsule network with dynamic routing for linear time NMT. \citet{yang-etal-2018-investigating} explored capsule networks for text classification with strategies to stabilize the dynamic routing process. \citet{gu2019improving} introduces capsule networks into Transformer to model the relations between different heads in multi-head attention. We specifically investigated dynamic routing algorithms for the document-level NMT.

\section{Conclusion}
We have proposed a novel Query-guided Capsule Network with an improved dynamic routing algorithm for enhancing context modeling for the document-level Neural Machine Translation Model. Experiments on English-German in different domains showed our model significantly outperforms sentence-level NMTs and achieved state-of-the-art performance on two of three datasets, which proved the effectiveness of our approaches.

\section*{Acknowledgments}
We thank the anonymous reviewers for their insightful comments. This work was supported by the National Natural Science Foundation of China (NO.61662077, NO.61876174) and National Key R\&D Program of China (NO.2017YFE9132900).

\bibliography{main}
\bibliographystyle{acl_natbib}

\end{document}